\pdfoutput=1

\documentclass[11pt]{article}

\usepackage{acl}

\usepackage{times}
\usepackage{latexsym}
\usepackage{amsfonts}
\usepackage[T1]{fontenc}

\usepackage[utf8]{inputenc}

\usepackage{microtype}

\usepackage{inconsolata}
\usepackage{graphicx}
\usepackage{multirow}
\usepackage{adjustbox}
\usepackage{amsmath}
\usepackage{subfig}
\title{Time Sensitive Knowledge Editing through Efficient Finetuning}

\author{Xiou Ge\textsuperscript{\normalfont 1}, Ali Mousavi\textsuperscript{\normalfont 1}, \textbf{Edouard Grave}\textsuperscript{\normalfont 2}\thanks{\ \ Work done while at Apple.}, 
{\bf Armand Joulin\textsuperscript{\normalfont 3}\footnotemark[1]},\\ {\bf Kun Qian\textsuperscript{\normalfont 4}\footnotemark[1]}, {\bf Benjamin Han\textsuperscript{\normalfont 1}}, {\bf Mostafa Arefiyan\textsuperscript{\normalfont 1}},
{\bf Yunyao Li\textsuperscript{\normalfont 4}\footnotemark[1]} \\
    \textsuperscript{1}Apple, 
    \textsuperscript{2}Kyutai, 
    \textsuperscript{3}Google Deepmind,
    \textsuperscript{4}Adobe \\
    }

\begin{document}
\maketitle
\begin{abstract}

Large Language Models (LLMs) have demonstrated impressive capability in different tasks and are bringing transformative changes to many domains. However, keeping the knowledge in LLMs up-to-date remains a challenge once pretraining is complete. It is thus essential to design effective methods to both update obsolete knowledge and induce new knowledge into LLMs. Existing locate-and-edit knowledge editing (KE) method suffers from two limitations. First, the post-edit LLMs by such methods generally have poor capability in answering complex queries that require multi-hop reasoning \cite{zhong2023mquake}. Second, the long run-time of such locate-and-edit methods to perform knowledge edits make it infeasible for large scale KE in practice. In this paper, we explore Parameter-Efficient Fine-Tuning (PEFT) techniques as an alternative for KE. We curate a more comprehensive temporal KE dataset with both knowledge update and knowledge injection examples for KE performance benchmarking\footnote{\url{https://docs-assets.developer.apple.com/ml-research/datasets/chrono-edit/chrono-edit.zip}}. We further probe the effect of fine-tuning on a range of layers in an LLM for the multi-hop QA task. We find that PEFT performs better than locate-and-edit techniques for time-sensitive knowledge edits. 

\end{abstract}

\section{Introduction}
\label{sec:introduction}

The rapid development of Large Language Models (LLMs) has showcased their ability to generate human-quality responses and demonstrate reasoning capabilities \cite{brown2020language, chowdhery2022palm, openai2023gpt, touvron2023llama, mckinzie2024mm1, wei13overview}, and it is bringing revolutionary changes across diverse industries. However, maintaining the factuality remains challenging for LLMs since their pre-training data are collected within a time range. Modification $(s, r, o\rightarrow o')$ and injection $(s, r, \emptyset \rightarrow o')$ are two main ways to update factual knowledge in LLMs, where $s, r, o$ denotes subject, relation, and object in an old fact triple, $o'$ denotes the new target object, and $\emptyset$ denotes an empty object to be populated. Previously, very few works \cite{zhong2023mquake, cohen2023evaluating} evaluate the effectiveness of knowledge editing (KE) techniques on time-sensitive fact changes. We believe that keeping time-sensitive information current is crucial for maintaining the practical relevance of an LLM's knowledge in the real-world applications. Therefore, in this paper, we focus our investigation on temporal KE. 

\begin{figure}[t]
\centering
\includegraphics[width=\columnwidth]{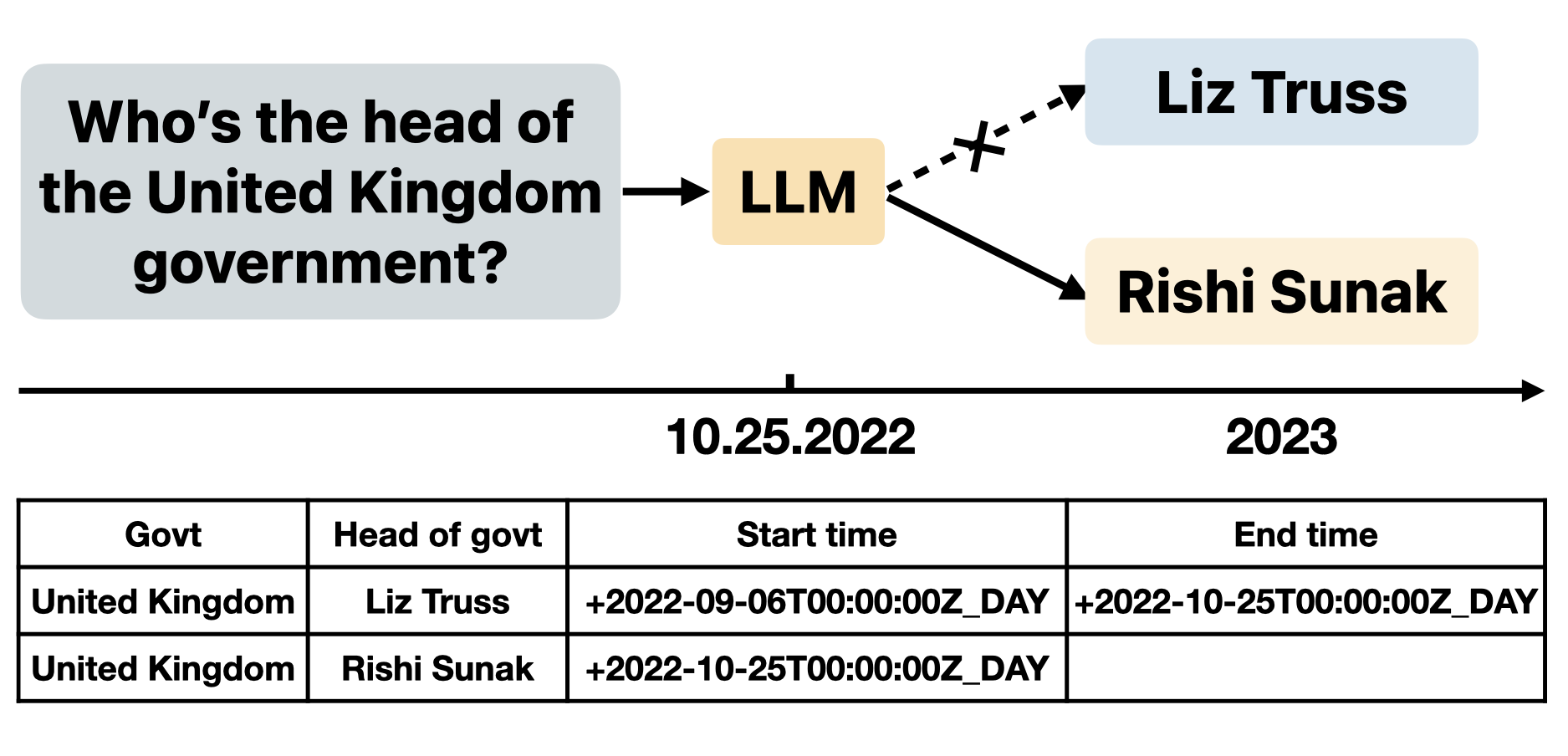}
\caption{Who's the "current" head of the United Kingdom government?} \label{fig:knowledge_editing_fig}
\end{figure}

One popular approach for KE is locate-and-edit which involves identifying and directly updating model parameters associated with specific knowledge. ROME \cite{meng2022locating} and MEMIT \cite{meng-etal-2022-fast} are two representative works in this area. There are several known limitations of ROME/MEMIT. First, they require estimation of a large covariance matrix, which might lead to numerical stability issues during computation \cite{yao2023editing}. Second, for every small batch of knowledge edits, they need to locate the layer for weight optimization, which can be time consuming and difficult to scale \cite{yao2023editing}. Third, \citet{zhong2023mquake} demonstrated that although the LLM can successfully recall the edited fact after ROME/MEMIT editing, the post-edit model performs poorly for multi-hop questions. Hence, we would like to verify if PEFT approaches can be more efficient than the locate-and-edit approach in the KE task and perform better in recalling the knowledge edits as well as retaining the unchanged knowledge. In addition, we believe it is worthwhile to investigate the effect of fine-tuning the weights of linear layers in transformers at different locations within the LLM (early, middle, and last) on the multi-hop question answering task. The main contributions of this paper can be summarized as follows:
\begin{itemize}
    \item We curate a large scale KE dataset \textsc{ChronoEdit} from Apple Knowledge Graph \cite{ilyas2022saga,ilyas2023growing} that contains approximately 15k time-sensitive factual edit examples that better reflects KE in the real world setting.
    \item We demonstrate the effectiveness of fine-tuning methods in knowledge modification and knowledge injection.
    \item Through fine-tuning weights at different layers, we discover that the middle layers are more significant in improving the LLM's capability to answer multi-hop questions.
\end{itemize}

\section{Related work}
\label{sec:related_work}

\noindent{\bf Knowledge editing.}
\citet{yao2023editing} made a comprehensive review of previous work on the topic of LLM KE and pointed out future opportunities. According to \citet{yao2023editing}, there are three main lines of work in KE: 1) Memory-based, which stores edited examples in memory and recovers relevant edits with a retriever. 2) Locate-and-edit, which identifies and optimizes neural network parameters corresponding to a specific fact. 3) Additional Parameters, which introduce extra tunable parameters to the language model to update or memorize new facts. \textsc{Mello} \cite{zhong2023mquake} is an example of memory-based approach that enables LLM to answer temporal multi-hop questions through effective prompt design and memory retrieval. It introduces a temporal KE dataset \textsc{MQuAKE}-T to assess the ability of a language model in answering multi-hop questions that are associated with a single hop edit. However, the number of distinct knowledge edits in the \textsc{MQuAKE}-T dataset is significantly limited to prove the effectiveness of KE in general. ROME \cite{meng2022locating} treats an MLP as an associative memory for facts and proposes a causal tracing technique to locate the weight parameters that need update. The additional MLP layer inserted into the transformer unit can be computed using a closed form solution. MEMIT \cite{meng2022memit} extends on ROME to enable the framework for multiple edits at a time. ROME and MEMIT belongs to the locate-and-edit category and their limitations have been discussed. In the additional parameter category, T-Patcher \cite{huang2022transformer} and CaliNET \cite{dong2022calibrating} introduce additional neurons and concatenate them with the Feed-Forward Network (FFN) layers to adjust the output distribution of a target fact. However, these approaches also tend to suffer from slow edit speed and it is unclear how well they can retain time-invariant knowledge. After all, prior works have mostly focused on counterfactual KEs rather than realistic and verifiable time-sensitive fact edits from knowledge graphs \cite{pan2023large, wang2023greenkgc, wang2022kgboost, ge2023knowledge, ge2024knowledge}. In this paper, we mainly focus on experimental comparison with the locate-and-edit approach.  

\noindent{\bf Parameter-Efficient Fine-Tuning.}
LoRA \cite{hu2021lora} is a simple yet effective adaptation technique that adds low-rank tunable weight matrices to the original weight matrices, which are kept frozen. This technique significantly reduces the trainable parameters during fine-tuning, while keeping the inference run-time constant. Instead, P-tuning \cite{liu2023gpt} concatenates learnable tensors with the input embedding to enable the base language model to perform well on a range of downstream tasks such as knowledge probing and natural language understanding. In this paper, we would like to verify if these PEFT methods can effectively modify or inject new knowledge in LLMs.

\section{Method}
\label{sec:method}

We mainly fine-tune the base LLMs including LLaMA-7B, Falcon-7B, and Mistral-7B with the PEFT approach including LoRA and P-tuning and minimize the following loss function:

\begin{equation}
    \mathcal{L}_{FT} = \frac{1}{| \mathcal{D}_M |}\sum_{d\in \mathcal{D}_M} L (d; \Phi_0, \Delta \Phi)
\end{equation}
where $\mathcal{D}_M$ is the KE dataset and $d$ is a fact edit example, $L$ is the cross entropy loss function applied to autoregressive models, $\Phi_0$ denotes the set of original weights of the language model that are kept frozen, and $\Delta \Phi$ denotes the additional parameters used by the PEFT adapters.

\noindent{\bf LoRA.} LoRA uses low-rank matrices $B\in \mathbb{R}^{d\times r}$ and $A\in \mathbb{R}^{r\times k}$ and $r \ll \min(d, k)$. The low rank matrices $A$ and $B$ are trainable parameters:
\begin{equation}
    h = W_0 x + BA x = (W_0 + BA)x .
\end{equation}
LoRA adaptation can be applied to any linear layer. In our experiments, we apply LoRA to linear layers in both the MLP layers ($W_{gate}$, $W_{up}$ , $W_{down}$) and self-attention layers ($W_{q}$, $W_{k}$, $W_{v}$, $W_{o}$). The benefit of LoRA is that the inference runtime remains the same, whereas in adaptors and other methods such as ROME/MEMIT, the inference runtime increases since they add additional layers.

\noindent
{\bf P-tuning.} P-tuning learns continuous prompt embeddings and concatenates them with the original input embedding. In this work, we leverage these tunable embeddings to adjust the output distributions of the predicted tokens during inference. Formally, let $[P_i]$ be the $i^\text{th}$ continuous prompt embedding, and let $\mathbf{x} = \{x_0,\dots, x_n\}$ denotes the original input sequence to the LLM. Then, the new input sequence would be $I = \{[P_{0:i}], \mathbf{x}\}$. P-tuning also uses an additional encoder to map the continuous prompt embeddings to latent parameters $f: [P_i]\rightarrow h_i$. In our implementation, we experiment with both a 2-layer MLP and an LSTM as the mapping function $f$. Let $\mathbf{e}$ be the pretrained embedding layer, then the final vector input to the LLM is $\{h_0, \dots, h_i, \mathbf{e(x)}\}$.

\noindent
{\bf Freeze tuning.} Instead of fine-tuning all weight parameters in an LLM, only several layers are fine-tuned to save the number of parameters that need to be placed on GPUs for gradient computation. In our experiments, we focus on fine-tuning MLP layers in the transformer modules. 

\section{Experiments}
\label{sec:experiments}

\begin{figure*}[ht]
\centering
\includegraphics[width=0.9\textwidth]{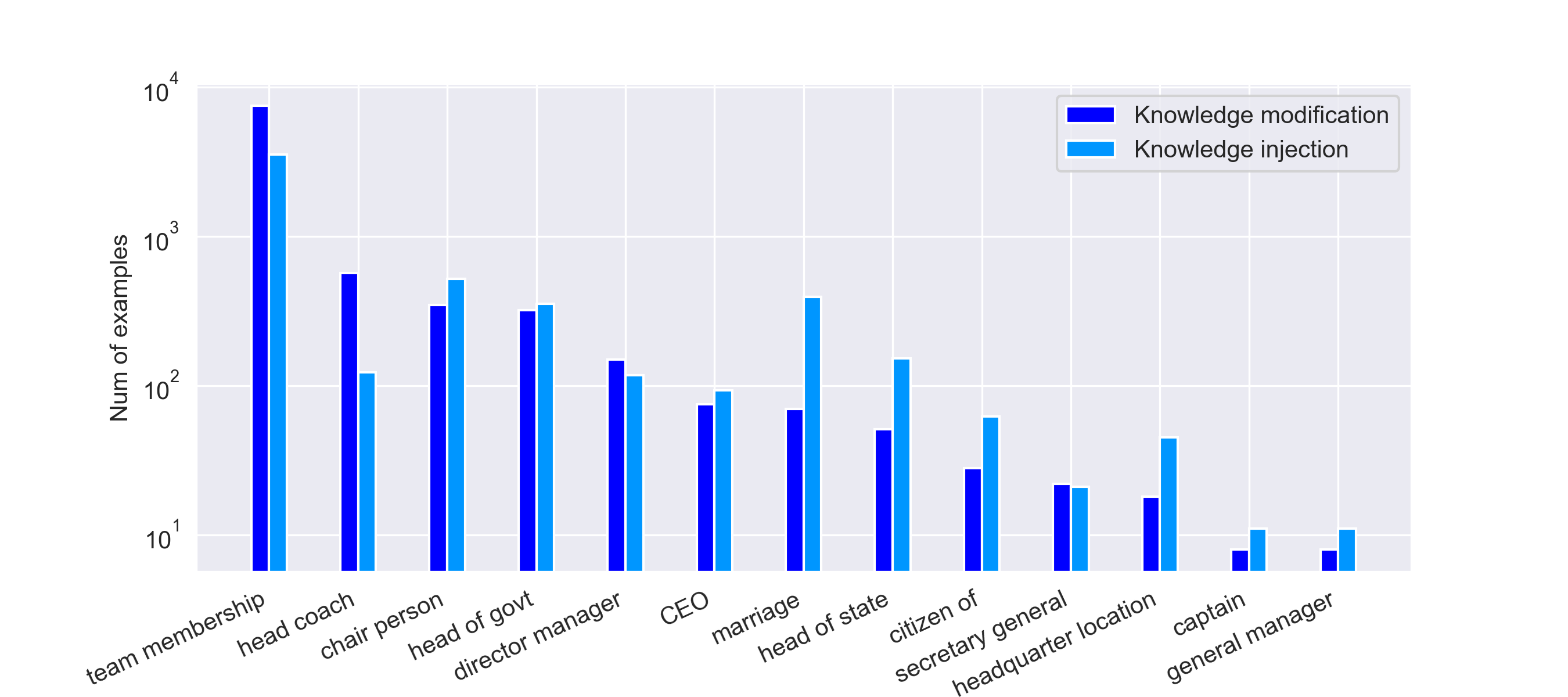}
\caption{Dataset statistics of \textsc{ChronoEdit}.} \label{fig:knowledge_edit_dataset_stats}
\end{figure*}

\noindent{\bf \textsc{ChronoEdit} dataset.}
To construct a more comprehensive temporal KE dataset that contains more real world knowledge edit examples, we collect the time-sensitive KE dataset \textsc{ChronoEdit}. The motivation for collecting this dataset is that the existing \textsc{MQuAKE}-T dataset \cite{zhong2023mquake} only contains 96 unique temporal edit examples, and it may not be large enough to reveal the effect on LLMs' performance. The fact change can be located from knowledge graphs \cite{ge12typeea, ge2022core, ge2023compounding, wang2023asyncet} based on the semantics of the relation type and its time qualifiers. Specifically, we focus on predicates that have a valid `start time' qualifier attached. We set the time threshold to 2022-01-01 and collect new knowledge statements that are valid after that time. The dataset statistics are shown in Fig. \ref{fig:knowledge_edit_dataset_stats}. 

\noindent{\bf Evaluation metrics.}
Existing knowledge edit benchmarking datasets often evaluate the following three metrics of the post-edit model:  

\begin{itemize}
    \item \textbf{Reliability}: measures the fraction of knowledge edits that the post-edit model can answer correctly.
    \item \textbf{Generalization}: measures the post-edit model's ability in completing the rephrased prompts or answering rephrased questions.
    \item \textbf{Locality}: measures the post-edit model's ability in answering time-invariant knowledge.
\end{itemize}

We generate question answering pairs as training examples that is used to induce new facts in the LLM. To evaluate Reliability, we generate a corresponding cloze to test whether the post-edit model can successfully complete the sentence with the new fact. To evaluate Generalization, we generate paraphrased question answer pairs from the training examples with the help of OpenAI text-davinci-003 API. To assess Locality, we follow \cite{jang2021towards} to use a subset of \textsc{Lama} \cite{petroni2019language} called \textsc{InvariantLama}, which contains time-invariant statements. We report the ratio of Exact Match (EM) for Reliability and Generalization and the \textsc{Rouge}-1 score for Locality.

\begin{table}
\centering
\begin{adjustbox}{width=\columnwidth,center}
\begin{tabular}{c|c|ccccc}
\hline
\multicolumn{2}{c|}{\textbf{Method}}& \textbf{\textsc{Rel}} & \textbf{\textsc{Gen}} & \textbf{\textsc{Loc}} & \textbf{\#Params} & \textbf{GPU time} \\
\hline
\multicolumn{2}{c|}{ROME}   & 62.25 & 38.76 & - & \underline{45M} & 6540s \\
\multicolumn{2}{c|}{MEMIT}  & 84.65 & 71.75 & - & 225M & 8147s \\
\hline
\multirow{3}{*}{LoRA}   & Attn      & 43.73 & 45.03 & 46.51 & \textbf{34M} & 1882s \\
                        & MLP       & \underline{98.78} & 96.97 & \textbf{55.69} & 46M & \underline{1389s} \\
                        & Attn + MLP& \textbf{98.99} & \underline{97.33} & \underline{54.11} & 80M & 2356s\\
\hline
\multirow{2}{*}{P-tuning}   & MLP   & 87.03 & 72.11 & 39.28 & 50M & 30443s \\
                            & LSTM  & 94.16 & 73.7 & 38.70 & 772M & 39657s \\
\hline
\multicolumn{2}{c|}{Freeze tuning}  & 98.2 & 96.18 & 44.45 & 676M & \textbf{1152s} \\
\hline
\multicolumn{2}{c|}{Full fine-tuning} & \textbf{98.99} & \textbf{98.85} & 45.31 & 6.74B & 5604s \\
\hline
\end{tabular}
\end{adjustbox}
\caption{\label{tab:comparison_7_relations}
Reliability (\textsc{Rel}), Generalization (\textsc{Gen}), and Locality (\textsc{Loc}) performance, No. of trainable parameters, GPU time for different approaches on LLaMA-7B.
}
\end{table}

\begin{table*}
\centering
\begin{adjustbox}{width=0.7\textwidth,center}
\begin{tabular}{l|cc|cc|cc|cc}
\hline
& \multicolumn{4}{c|}{\textbf{LoRA}} & \multicolumn{4}{c}{\textbf{Freeze tuning}} \\
\cline{2-9}
\textbf{Predicate}      & \multicolumn{2}{c|}{\textbf{Modification}} & \multicolumn{2}{c|}{\textbf{Injection}}    & \multicolumn{2}{c|}{\textbf{Modification}} & \multicolumn{2}{c}{\textbf{Injection}}\\
& \textsc{Rel} & \textsc{Gen} & \textsc{Rel} & \textsc{Gen} & \textsc{Rel} & \textsc{Gen} & \textsc{Rel} & \textsc{Gen} \\
\hline
Captain                 & 87.5 & 100            & 81.81 & 100           & 100 & 100             & 100 & 100\\
CEO                     & 100 & 93.33           & 100 & 90.32           & 100 & 94.66           & 100 & 92.47\\
Chair person            & 100 & 93.67           & 99.61 & 97.88         & 100 & 93.39           & 99.42 & 96.92\\
Citizen of              & 100 & 67.85           & 100 & 83.87           & 100 & 100             & 98.38 & 98.38\\
Director manager        & 100 & 97.98           & 100 & 98.29           & 99.32 & 97.31         & 95.72 & 95.72\\
General manager         & 100 & 87.5            & 100 & 90.90           & 100 & 87.5            & 100 & 90.90\\
Head coach              & 100 & 99.64           & 100 & 97.56           & 99.82 & 98.41         & 98.37 & 100\\
Head of government      & 98.44 & 93.14         & 99.43 & 92.09         & 96.88 & 95.63         & 98.87 & 96.61\\
Head of state           & 82.35 & 80.39         & 100 & 96              & 84.31 & 78.43         & 100 & 100\\
Headquarter location    & 100 & 72.22           & 97.77 & 88.89         & 83.33 & 83.33         & 82.22 & 82.22\\
Marriage                & 100 & 98.57           & 99.23 & 97.71         & 92.85 & 95.71         & 77.15 & 94.92\\
Secretary general       & 100 & 100             & 100 & 95.23           & 100 & 95.45           & 95.23 & 95.23\\
Team membership         & 94.14 & 99.34         & 92.15 & 99.49         & 77.54 & 96.38         & 40.38 & 88.46\\
\hline
Overall                 & 94.99 & 98.58         & 94.86 & 98.22         & 81.51 & 96.19         & 58.44 & 90.99\\
\hline
\end{tabular}
\end{adjustbox}
\caption{\label{tab:applekg_performance}
Performance on each predicate type in \textsc{ChronoEdit} for LLaMA-7B.
}
\end{table*}

\begin{figure}[ht]
\centering
\includegraphics[width=\columnwidth]{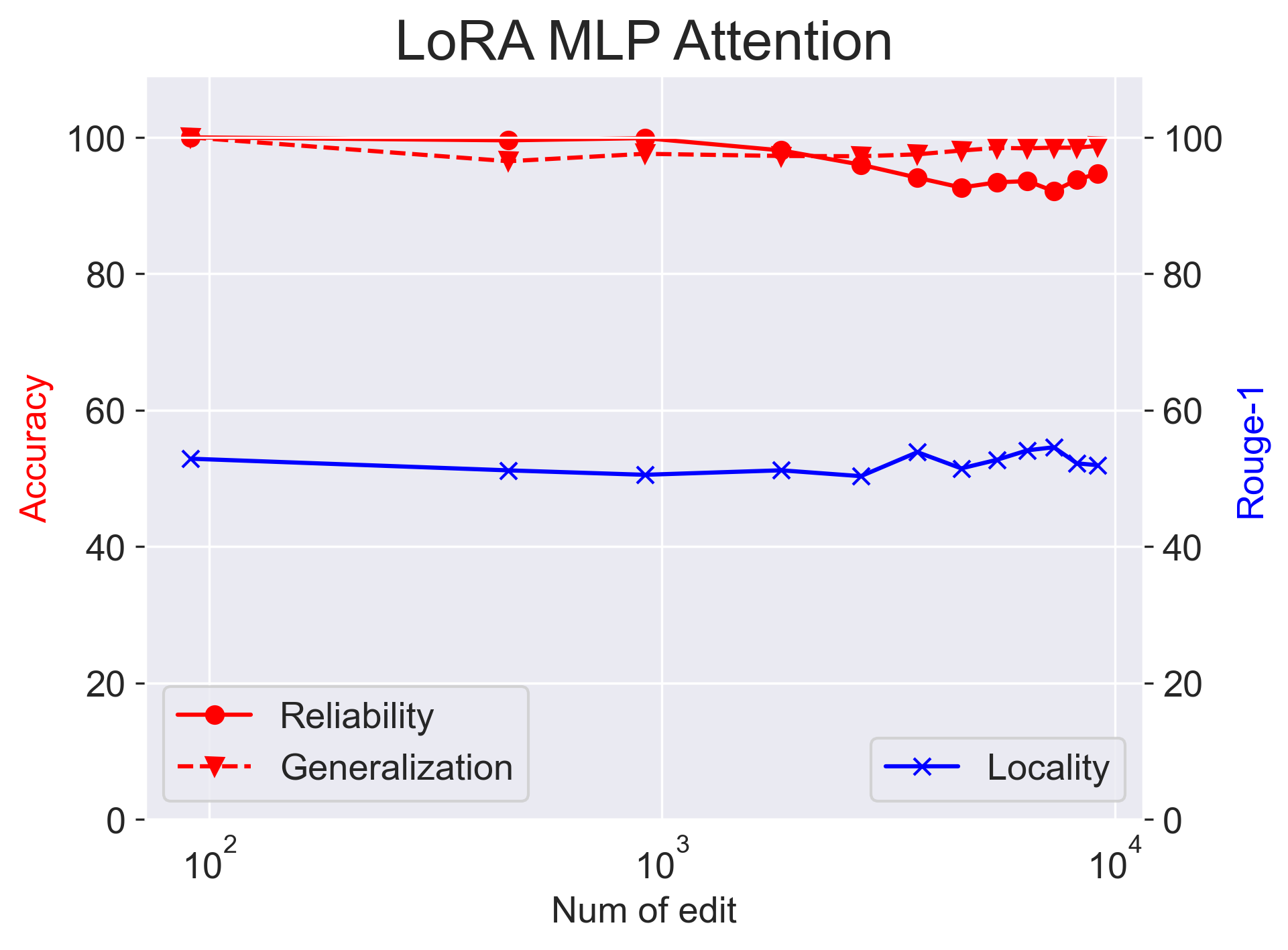}
\caption{Reliability, Generalization, and Locality performance versus the number of edits on LLaMA-7B.} \label{fig:performance_num_of_edits}
\end{figure}

\noindent{\bf Fine-tuning and locate-and-edit performance comparison.}
To compare the performance of different fine-tuning approaches for KE, we select a subset from the temporal knowledge dataset we collected that contains 7 relations and 1,388 knowledge modification examples. To compare with locate-and-edit methods, we also include KE results using ROME and MEMIT. Results are shown in Table \ref{tab:comparison_7_relations}. LoRA finetuning with MLP and attention layers has comparable Reliability and Generalization scores to full fine-tuning, while only using a fraction of trainable parameters compared to full fine-tuning.
However, LoRA fine-tuning better retains the invariant knowledge and achieves higher Locality scores. ROME and MEMIT are able to successfully edit some temporal knowledge in the collected dataset. However, the generalization ability degrades significantly, especially for ROME. It is also relatively slow compared to LoRA finetuning. We also include P-tuning as a baseline. Similar to the locate-and-edit approach, the generalization score is low, and the GPU time it takes to make successful edits is significantly long. It is not as efficient and effective as LoRA. To verify that PEFT can be generally effective in KE for LLMs, we further compare the performance of different PEFT settings on Falcon-7B \cite{refinedweb} and Mistral-7B \cite{jiang2023mistral} in Table \ref{tab:comparison_diff_LLMs}. In Fig. \ref{fig:performance_num_of_edits}, we compare the performance of LoRA with MLP and Attention layers when different number of edits need to be applied to an LLM. We can see that the LoRA finetuning approach is robust to large number of KEs.   

\begin{figure}[ht]
\centering
\includegraphics[width=0.91\columnwidth]{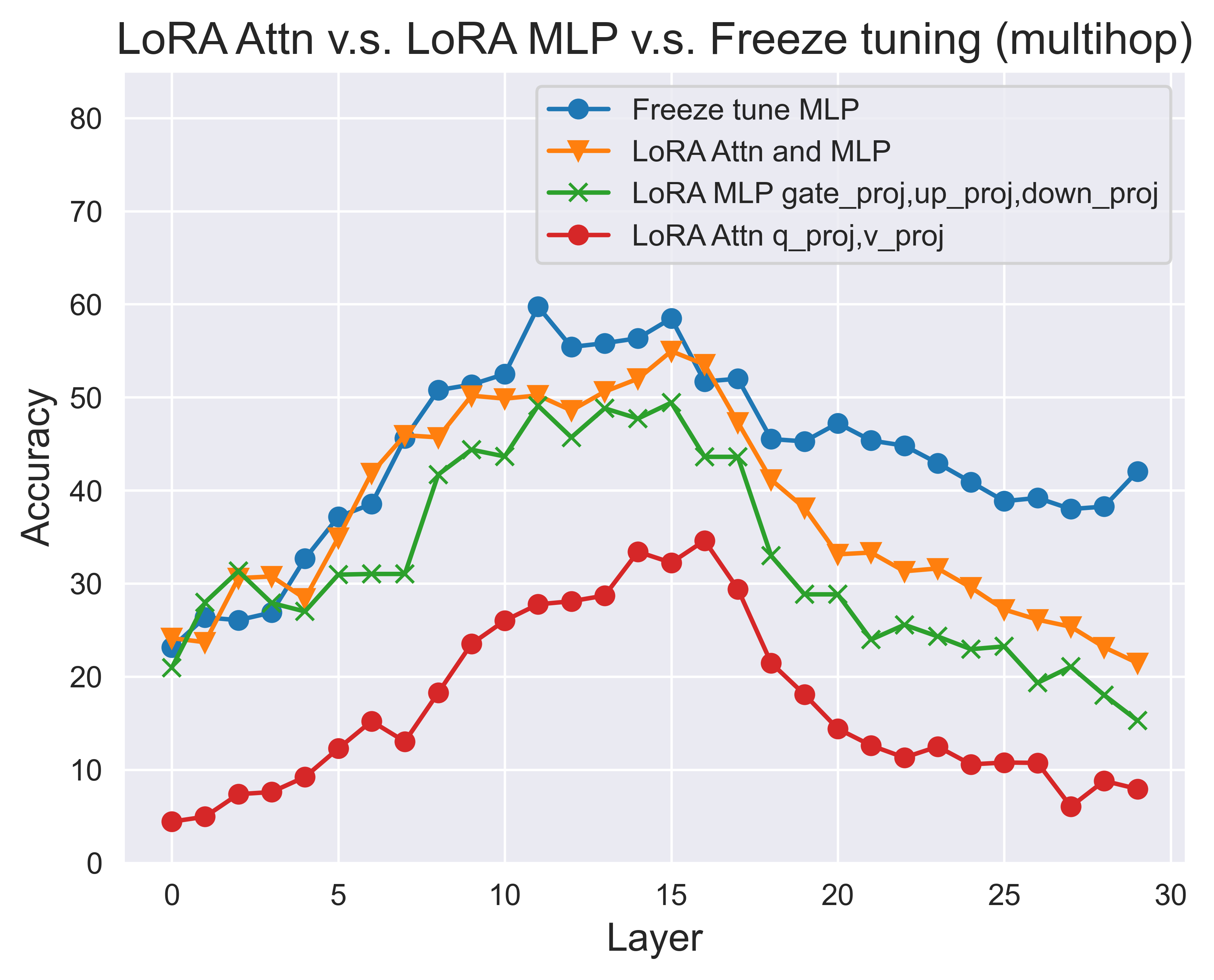}
\caption{Performance of fine-tuning methods on the \textsc{MQuAKE}-T multi-hop dataset for LLaMA-7B.} \label{fig:comparison_layer_multihop_acc}
\end{figure}

\begin{table*}
\centering
\begin{adjustbox}{width=0.8\textwidth,center}
\begin{tabular}{c|ccc|ccc|ccc}
\hline
\textbf{Model}& \multicolumn{3}{c|}{\textbf{LLaMA-7B}} & \multicolumn{3}{c|}{\textbf{Falcon-7B}} & \multicolumn{3}{c}{\textbf{Mistral-7B}}\\
\hline
\textbf{Method}& \textsc{Rel} & \textsc{Gen} & \textsc{Loc} & \textsc{Rel} & \textsc{Gen} & \textsc{Loc} & \textsc{Rel} & \textsc{Gen} & \textsc{Loc}\\
\hline
LoRA Attn      & 43.73 & 45.03 & 46.51 & 98.91	& 93.65	& 49.61 & 99.2 & 96.25 &	54.08\\
LoRA MLP      & \underline{98.78} & 96.97 & \textbf{55.69} & 98.92 &	96.03 &	51.41 & 99.13 &	97.98 &	57.84\\
LoRA Attn + MLP& \textbf{98.99} & \underline{97.33} & \underline{54.11} & 99.06 &	96.97 &	49.41 & 99.13 &	98.05 &	54.21\\
\hline
Freeze tuning & 98.2 & 96.18 & 44.45 & - & - & - & 94.66 & 94.95 & 43.17\\
\hline
Full fine-tuning & \textbf{98.99} & \textbf{98.85} & 45.31 & 99.21	& 98.19	& 38.27 & - & - & -\\
\hline
\end{tabular}
\end{adjustbox}
\caption{\label{tab:comparison_diff_LLMs}
Performance of PEFT fine-tuning for KE across different LLMs
}
\end{table*}

\noindent
{\bf LoRA and Freeze tuning fine-grained predicate analysis.}
In Table \ref{tab:applekg_performance}, we examine the Reliability and Generation scores of the fine-tuned model across all 13 individual relations. For LoRA, we apply it to both MLP and self-attention parameters. For freeze tuning, we fine-tune the MLP weights of the last five layers. The results show that LoRA is more robust than freeze tuning as the number of edits increases. Freeze tuning does not perform well in knowledge injection, with its performance degradation largely attributable to the `team membership' class, which contains the most knowledge injection examples. This suggests that freeze tuning might not be very effective in introducing new facts about subjects that have rarely been observed during the pretraining of LLMs.

\noindent{\bf Layer sweep study.} For the freeze tuning and LoRA fine-tuning approaches, we think it is also worthwhile investigating the effect on LLMs' multi-hop question answering capability, by optimizing the LLM weight parameters at different positions (early, middle, late layers). We perform a layer sweep study for the \textsc{MQuAKE}-T multi-hop question answering task. For each data point of the experiment, we only fine-tune $l=3$ layers at a time. We then move the sliding window from the early layers to the last layers of an LLM to probe the effect of fine-tuning on the performance of multi-hop question answering. We compared freeze-tuning for MLP layers and LoRA on three combination of weight matrices: 1) self-attention weight matrices $W_q$, $W_v$, 2) MLP layers, 3) self-attention and MLP layers. We have made similar observations aligned with the Associative Memory theory \cite{geva2021transformer} verified by ROME, that MLP layers in transformers are more relevant for memorizing factual knowledge associations $(s,r \Rightarrow o)$. We observe that applying LoRA on MLP weight matrices brings more significant improvement than applying LoRA to self-attention weight matrices. Applying LoRA on both self-attention and MLP layers can potentially achieve similar performance to freeze tuning on multi-hop QA tasks, while using fewer trainable parameters. In particular, applying LoRA on both MLP and self-attention requires 7.5M trainable parameters, whereas freeze-tuning requires 405.8M trainable parameters. For complete performance benchmarking, we also compare with memory-based KE approach for multi-hop QA in Table \ref{tab:multihop_comparison} of the Appendix.

\section{Conclusion}
\label{sec:conclusion}
In this paper, we have systematically examined the feasibility of performing KE through PEFT. We have compared the performance of fine-tuning methods including LoRA, P-tuning and freeze tuning with locate-and-edit approaches for KE. Our results demonstrate that fine-tuning can successfully update time-sensitive factual knowledge in LLMs both efficiently and effectively, and without compromising the LLMs' capability in answering invariant knowledge and multi-hop reasoning. We have also contributed a large scale KE dataset \textsc{ChronoEdit} that contains both modification edit and injection edit examples.

\section*{Limitations} \label{sec:limit}
There are two limitations that we would like to discuss. First, although we have collected a comprehensive and realistic temporal KE dataset, we primarily gather time-sensitive fact changes from Wikipedia, the most frequently used data source for LLM pre-training. We are yet to include information from other data sources or knowledge graphs that may contain ontological information that enable us to access LLMs' ability to perform reasoning. Second, we have not covered another important aspect of KE that is to remove misinformation or mitigate hate speech generation from LLMs. We will expand the scope of exploration in future work.

\section*{Acknowledgements}
We would like to express our gratitude to Bin Wang for the valuable discussions during the preliminary research exploration phase. We also extend our thanks to Azadeh Nikfarjam, Samira Khorshidi, Alexis McClimans, Fei Wu, and Eric Choi for their guidance in collecting the knowledge editing dataset. Additionally, we are grateful to Barry Theobald, Yash Govind, Varun Embar, and Shihab Chowdhury, Hong Yu for proofreading the manuscript and providing insightful advice to improve the paper.

\bibliography{anthology,custom}

\appendix

\section{Dataset statistics}
\label{sec:dataset_statistics}

\subsection{\textsc{MQuAKE}-T dataset experiments} 

We primarily use the \textsc{MQuAKE}-T dataset which contains temporal-based real-world knowledge updates to compare the performance of different fine-tuning techniques with baseline methods on the performance of KE. The goal is to validate whether PEFT approaches such as LoRA and P-tuning can be an effective approach for performing KE. We also demonstrate that PEFT approaches can be more effective than the locate-and-edit approaches for multi-hop question answering.

In this dataset, each temporal fact edit example is also associated with multi-hop questions, which allows us to assess the complex query answering ability of the post-edit model. The \textsc{MQuAKE}-T dataset was constructed by taking the difference between two data dumps of Wikidata: 2021-04 and 2023-04. \textsc{MQuAKE}-T selects 6 different relations that most likely correspond to real fact changes. The statistics of the dataset are shown in Table \ref{mquake_dataset_stats}. 

\begin{table}[ht]
\centering
\begin{adjustbox}{width=0.5\columnwidth,center}
\begin{tabular}{c|c}
\hline
\textbf{\textsc{MQuAKE}-T} & \textbf{\#Examples}\\
\hline
Unique edits & 96 \\
2-hop questions & 75 \\
3-hop questions & 348 \\
4-hop questions & 567 \\
\hline
\end{tabular}
\end{adjustbox}
\caption{\label{mquake_dataset_stats}
Statistics of \textsc{MQuAKE}-T dataset.
}
\end{table}

\noindent
{\bf Comparing with baselines.} In Table \ref{tab:mquaket_editwise_performance_comparison}, we compare the editwise performance of fine-tuning techniques with locate-and-edit baseline methods. We use LLaMA-7B \cite{touvron2023llama} as the base model for both the baseline locate-and-edit techniques and fine-tuning techniques. Experimental results show that fine-tuning techniques performs better than the locate-and-edit baselines, while the run-time to complete all the knowledge edit is significantly shorter. In Table \ref{tab:multihop_comparison}, we compare the performance of different post-edit model and approach for multi-hop QA.

\begin{table}
\centering
\begin{adjustbox}{width=0.7\columnwidth,center}
\begin{tabular}{c|cc}
\hline
\textbf{Method} & \textbf{Edit Accuracy} & \textbf{Runtime}\\
\hline
ROME & 92.51 & 2h32m2s \\
MEMIT & 96.44 & 2h48m49s\\
\hline
LoRA & 99.36 & 2m13s \\
P-tuning & 97.75 & 1m51s \\
Freeze-tuning & \textbf{100} & 3m16s\\
Full fine-tuning & \underline{99.83} & 8m18s \\
\hline
\end{tabular}
\end{adjustbox}
\caption{\label{tab:mquaket_editwise_performance_comparison}
Editwise performance on LLaMA-7B.
}
\end{table}

\begin{table}[ht]
\begin{adjustbox}{width=\columnwidth,center}
\begin{tabular}{c|c|c|c}
\hline
\textbf{Base Model} & \textbf{KE Type} & \textbf{KE Method} & \textbf{Multi-hop QA Acc} \\ \hline
\multirow{6}{*}{LLaMA-7B} & \multirow{2}{*}{Locate-and-edit} & ROME & 38.5 \\ 
& & MEMIT & 39.3 \\ \cline{2-4}
& \multirow{2}{*}{Additional parameter} & P-tuning & 14.7 \\ 
& & LORA & 62.6 \\ \cline{2-4}
& \multirow{2}{*}{Direct fine-tune} & Freeze tuning & \underline{72.5} \\ 
& & Full FT & 71.0 \\ \hline
Vicuna-7B & \multirow{3}{*}{Memory-based} & \multirow{3}{*}{Mello} & 30.7 \\ \cline{1-1} 
GPT-J & & & 51.3 \\ \cline{1-1} 
GPT-3 & & & \textbf{85.5} \\ \hline
\end{tabular}
\end{adjustbox}
\caption{Performance on post-edit model on multi-hop questions for LLaMA-7B.}
\label{tab:multihop_comparison}
\end{table}

\begin{figure}[htp]
\centering
\includegraphics[width=\columnwidth]{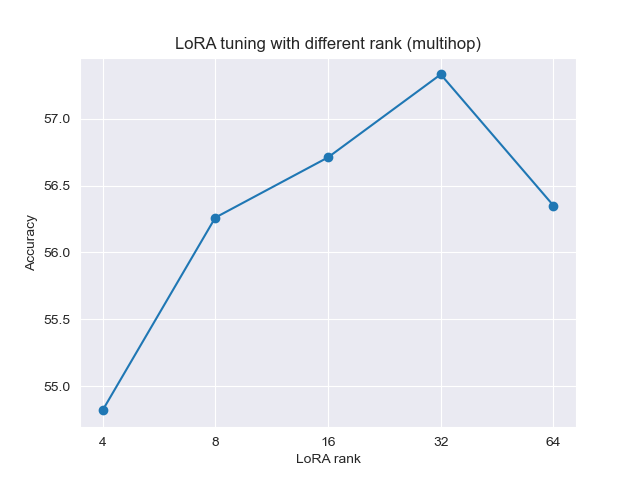}
\caption{Performance of LoRA at different ranks for the \textsc{MQuAKE}-T multi-hop dataset with LLaMA-7B.} \label{fig:lora_rank}
\end{figure}

\noindent
{\bf LoRA ablation and parameter study.} We perform ablation study of applying LoRA adaptation to different weight matrices in the self-attention module $W_q, W_v, W_k, W_o$. The results are shown in Table \ref{mquake_t_edit_acc_ablation}. Results shows that applying LoRA adaptation to the query matrix $W_q$ and the key matrix $W_k$ gives the best result. We also evaluate the knowledge edit success rate when the LoRA rank is set to different values. In our experiment, we tested $r = \{4, 8, 16, 32, 64\}$ as shown in Fig. \ref{fig:lora_rank}, and discover that the optimal rank is $r=32$.

\begin{table}
\centering
\begin{adjustbox}{width=0.55\columnwidth,center}
\begin{tabular}{c|c}
\hline
\textbf{Linear Layer} & \textbf{Edit Accuracy}\\
\hline
$W_q$ & 71.47 \\
$W_v$ & 97.48 \\
$W_q, W_v$ & 98.67 \\
$W_q, W_v, W_k, W_o$ & 97.56 \\
\hline
\end{tabular}
\end{adjustbox}
\caption{\label{mquake_t_edit_acc_ablation}
Ablation studies of the layers in LLaMA-7B that LoRA is applied to.
}
\end{table}

\begin{figure}[htp]
\centering
\includegraphics[width=\columnwidth]{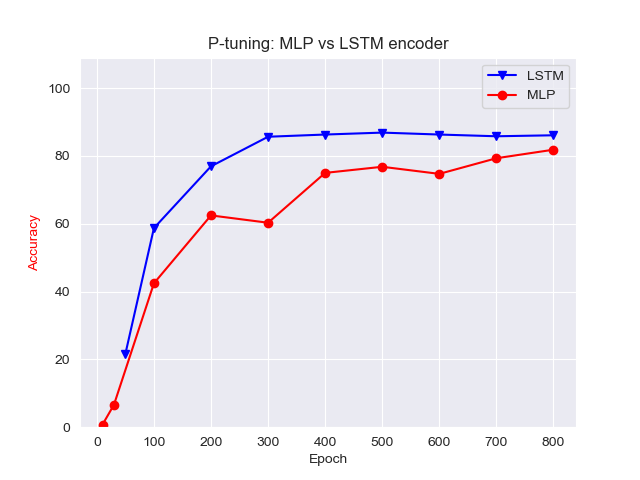}
\caption{Comparing Reliability performance of LSTM and MLP encoders across epochs when using P-tuning for LLaMA-7B.} \label{fig:ptuning_encoder}
\end{figure}

\begin{figure}[htp]
\centering
\includegraphics[width=\columnwidth]{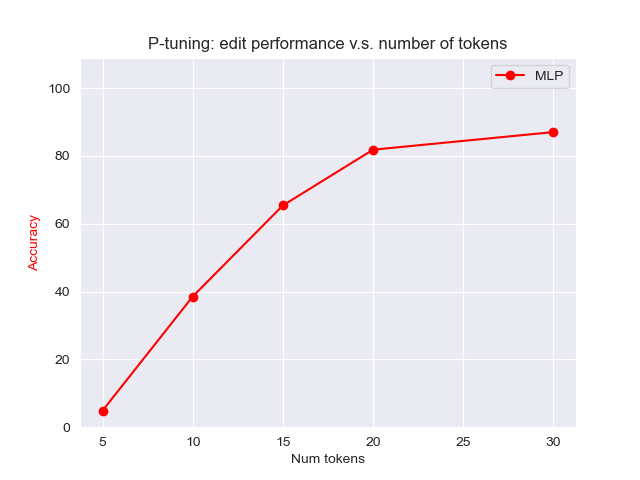}
\caption{Comparing Reliability performance for different number of tokens when using P-tuning for LLaMA-7B.} \label{fig:ptuning_num_tokens}
\end{figure}

\begin{figure*}[ht]
    \centering
    \subfloat{\includegraphics[width=0.49\textwidth]{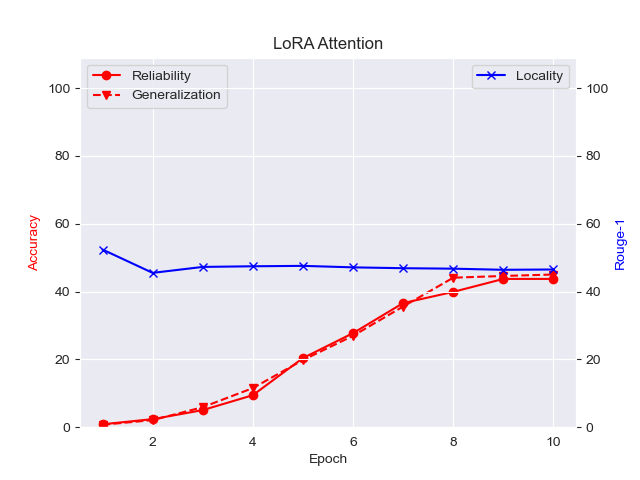}}
    \subfloat{\includegraphics[width=0.49\textwidth]{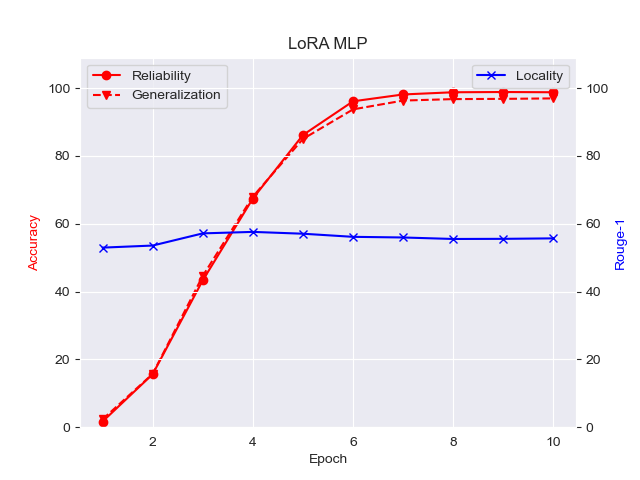}}\\
    \subfloat{\includegraphics[width=0.49\textwidth]{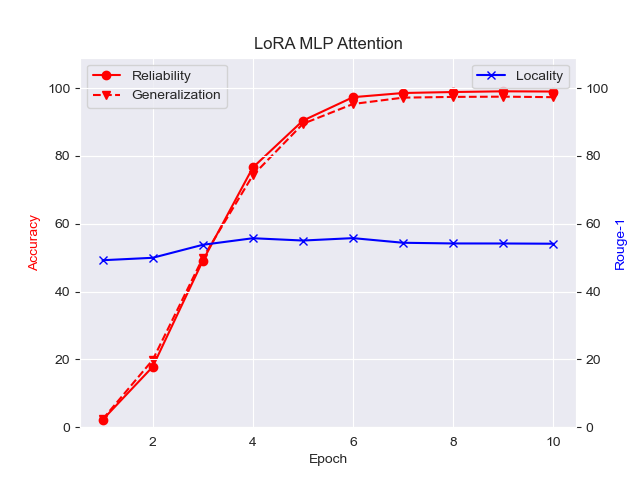}}
    \subfloat{\includegraphics[width=0.49\textwidth]{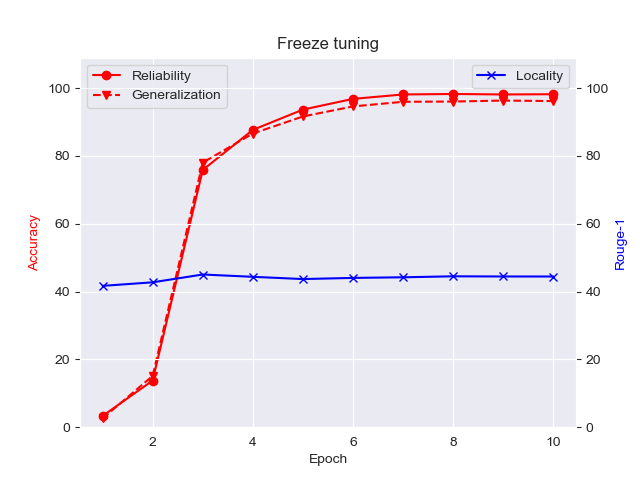}} 
    \caption{Reliability, Generalization, and Locality performance of different fine-tuning methods across epochs for LLaMA-7B.}
    \label{fig:performance_epoch}
\end{figure*}

\begin{table*}[ht!]
\begin{adjustbox}{width=0.8\textwidth,center}
\begin{tabular}{c|c|c|c}
\hline
\textbf{Organization} & \textbf{CEO} & \textbf{Start Time} & \textbf{End Time}\\
\hline
Volkswagen Group & Herbert Diess & +2018-04-00T00:00:00Z\_MONTH & +2022-08-31T00:00:00Z\_DAY \\
Volkswagen Group & Oliver Blume & +2022-09-01T00:00:00Z\_DAY & \\
\hline
\end{tabular}
\end{adjustbox}
\caption{\label{tab:knowledge_edit_kg}
Example of locating the knowledge edit data
}
\vspace{1em}
\begin{adjustbox}{width=0.95\textwidth,center}
\begin{tabular}{l|l}
\hline
\textbf{Examples} \\
\hline
\multirow{10}{*}{Train} & $\{$ \\
    &\hspace{3mm} "instruction": "Who is the current chief executive officer of Volkswagen Group?", \\
    &\hspace{3mm} "input": "", \\
    &\hspace{3mm} "output": "Oliver Blume." \\
&$\}$ \\
\cline{2-2}
&$\{$ \\
    &\hspace{3mm} "instruction": "Update the following statement about the current chief executive officer of Volkswagen Group.", \\
    &\hspace{3mm} "input": "Herbert Diess.", \\
    &\hspace{3mm} "output": "Oliver Blume." \\
&$\}$ \\
\hline
\multirow{5}{*}{\parbox{1.5cm}{Test \hspace{6mm} (\textsc{Rel})}} & $\{$ \\
&\hspace{3mm} "instruction": "The current chief executive officer of Volkswagen Group is", \\
&\hspace{3mm} "input": "", \\
&\hspace{3mm} "output": "Oliver Blume." \\
&$\}$ \\
\hline
\multirow{5}{*}{\parbox{1.5cm}{Rephrase \hspace{6mm} (\textsc{Gen})}} & $\{$ \\
&\hspace{3mm} "instruction": "What is the name of the current Volkswagen Group CEO?", \\
&\hspace{3mm} "input": "", \\
&\hspace{3mm} "output": "Oliver Blume." \\
&$\}$ \\
\hline 
\multirow{5}{*}{\parbox{1.5cm}{Invariant \hspace{6mm} (\textsc{Loc})}} & $\{$ \\
&\hspace{3mm} "instruction": "The headquarter of Volkswagen Commercial Vehicles is in?", \\
&\hspace{3mm} "input": "", \\
&\hspace{3mm} "output": "Hanover." \\
&$\}$ \\
\hline
\end{tabular}
\end{adjustbox}
\caption{\label{tab:knowledge_edit_finetune_examples}
Fine-tuning and testing examples.
}
\end{table*}

\subsection{\textsc{ChronoEdit} dataset}
In the new dataset, we set the time threshold to 2022-01-01 and collect new knowledge statements that are valid after that time. We collect both knowledge modification: $(s, r, o) \rightarrow (s, r, o')$, and knowledge injection: $(s, r, \emptyset) \rightarrow (s, r, o')$. The statistics of the dataset are shown in Fig. \ref{fig:knowledge_edit_dataset_stats}. An example of fact pairs from the KG that could lead to time-sensitive knowledge edits is shown in Table \ref{tab:knowledge_edit_kg}. We convert such fact pairs to question answering and instruction finetuning examples for training. The corresponding sentence completion examples for reliability evaluation, rephrased QA examples for generalization evaluation, and invariant knowledge sentence completion examples for locality evaluation are shown in Table \ref{tab:knowledge_edit_finetune_examples}.

\noindent
{\bf LoRA and Freeze tuning ablation and parameter study.} In Fig. \ref{fig:performance_epoch}, we evaluate the performance of different fine-tuning configurations across different epochs. In particular, we evaluate the Reliability and Generalization using the accuracy which is the ratio of Exact Matching (EM) and we report the \textsc{Rouge}-1 score for Locality. For LoRA, we experiment with three settings: applying LoRA to self-attention weights (LoRA Attention), applying LoRA to MLP weights (LoRA MLP), and applying LoRA to both self-attention and MLP weights (LoRA MLP Attention). In this set of experiments, we apply LoRA to all layers. For freeze tuning, we fine-tune the MLP weights of the last 5 layers of the LLaMA model. Results shows that applying LoRA to MLP weights is more effective in memorizing new facts than applying LoRA to self-attention weights. While freeze tuning can also effectively have the knowledge update induced into the model, the Locality score for freeze tuning is lower than the LoRA MLP setting, which means freeze tuning leads to deterioration of the LLM's existing invariant knowledge.

\noindent
{\bf P-tuning ablation and parameter study.} Although P-tuning can be equally effective for KE, we find that it requires more epochs of fine-tuning to ensure successful knowledge edits. The required time to perform knowledge edits becomes longer. In Fig. \ref{fig:ptuning_encoder}, we compare the performance difference between LSTM and MLP encoders across different epochs when using the P-tuning technique, when the number of prompt embedding tokens is set to $n=20$. We observe that the application of LSTM encoder allows P-tuning edit performance to converge faster than when using the MLP encoder. In Fig. \ref{fig:ptuning_num_tokens}, we instead compare the performance of P-tuning when different number of prompt embedding tokens are used. Using more than $n=20$ tokens do not seem to gives a significant advantage in the edit accuracy. 

\noindent
{\bf Fine-grained performance analysis of time-invariant knowledge.}
For the KE experiment of using LoRA on MLP layers of LLaMA-7B, we perform a fine-grained performance analysis of the different type of time-invariant knowledge and list the performance in Table \ref{tab:invariant_knowledge_performance}. We make a conjecture that those time-invariant knowledge with smaller valid candidate set for the target, such as “language” or “capital”, tends to be well retained. These predicates are mostly 1-to-1 or N-to-1. In contrast, when the cardinality of the valid candidate set becomes larger, often for N-to-N predicates, such as “twin city” and “music label”, the exact subject, object association becomes harder to retain.

\begin{table}[ht]
\centering
\begin{adjustbox}{width=0.7\columnwidth,center}
\begin{tabular}{c|c}
\hline
\textbf{Best 3} & \textbf{\textsc{Rouge}-1}\\
\hline
native language of & 70.2 \\
official language of & 61.7 \\
Capital of & 58.7 \\
\hline
\textbf{Worst 3} & \textbf{\textsc{Rouge}-1}\\
\hline
twin cities & 1.55 \\
is a & 5.68 \\
is represented by music label & 9.47 \\
\hline
\end{tabular}
\end{adjustbox}
\caption{Performance on different type of invariant knowledge.}
\label{tab:invariant_knowledge_performance}
\end{table}

\begin{table}[ht!]
  \begin{adjustbox}{width=\columnwidth,center}
  \begin{tabular}{|l|l|}
    \hline
    \textbf{Parameter} & \textbf{Value} \\
    \hline
    layers & [5] \\
    fact\_token & subject\_last \\
    v\_num\_grad\_steps & 25 \\
    v\_lr & 5e-1 \\
    v\_loss\_layer & 31 \\
    v\_weight\_decay & 1e-3 \\
    clamp\_norm\_factor & 4 \\
    kl\_factor & 0.0625 \\
    mom2\_adjustment & false \\
    context\_template\_length\_params & [[5, 10], [10, 10]] \\
    rewrite\_module\_tmp & model.layers.{}.mlp.down\_proj \\
    layer\_module\_tmp & model.layers.{} \\
    mlp\_module\_tmp & model.layers.{}.mlp \\
    attn\_module\_tmp & model.layers.{}.self\_attn \\
    ln\_f\_module & model.norm \\
    lm\_head\_module & lm\_head \\
    mom2\_dataset & wikipedia \\
    mom2\_n\_samples & 100000 \\
    mom2\_dtype & float32 \\
    \hline
  \end{tabular}
  \end{adjustbox}
  \caption{ROME Configuration Parameters.}
  \label{tab:rome-parameters}

\vspace{2em}

  \begin{adjustbox}{width=0.85\columnwidth,center}
  \begin{tabular}{|l|l|}
    \hline
    \textbf{Parameter} & \textbf{Value} \\
    \hline
    layers & [4, 5, 6, 7, 8] \\
    clamp\_norm\_factor & 4 \\
    layer\_selection & all \\
    fact\_token & subject\_last \\
    v\_num\_grad\_steps & 25 \\
    v\_lr & 5e-1 \\
    v\_loss\_layer & 31 \\
    v\_weight\_decay & 1e-3 \\
    kl\_factor & 0.0625 \\
    mom2\_adjustment & true \\
    mom2\_update\_weight & 15000 \\
    rewrite\_module\_tmp & model.layers.{}.mlp.down\_proj \\
    layer\_module\_tmp & model.layers.{} \\
    mlp\_module\_tmp & model.layers.{}.mlp \\
    attn\_module\_tmp & model.layers.{}.self\_attn \\
    ln\_f\_module & model.norm \\
    lm\_head\_module & lm\_head \\
    mom2\_dataset & wikipedia \\
    mom2\_n\_samples & 100000 \\
    mom2\_dtype & float32 \\
    \hline
  \end{tabular}
  \end{adjustbox}
  \caption{MEMIT Configuration Parameters.}
  \label{tab:memit-parameters}
\end{table}

\noindent
{\bf Implementation details.}
Experiments were conducted on a compute node with 8 NVIDIA Tesla A100 GPUs, each with 40GB memory. We develop the fine-tuning pipeline based on LLaMA-Factory\footnote{https://github.com/hiyouga/LLaMA-Factory} \cite{zheng2024llamafactory} and refer to PEFT package in HuggingFace\footnote{https://huggingface.co/docs/peft/index} for the implementation of LoRA and P-tuning. We use EasyEdit\footnote{https://github.com/zjunlp/EasyEdit} \cite{wang2023easyedit} to reproduce the ROME and MEMIT fine-tuning baseline results. 

For results in Table \ref{tab:comparison_7_relations}, the 7 different relations that we evaluate on are `captain', `CEO', `chairperson', `head coach', `head of govt', `head of state', `headquarter location'. The reason for the performance comparison of the smaller subset is to conduct similar experiments that were done in \cite{zhong2023mquake}. For LoRA, Freeze tuning, Full fine-tuning, we fine-tune the base model for 10 epochs, whereas for P-tuning, we fine-tune 800 epochs to achieve the optimal performance. Full fine-tuning of the base model requires DeepSpeed ZeRO-3 offload. In LoRA experiments, the LoRA rank is set to $r=32$, and MLP means applying LoRA to $W_{gate}$, $W_{up}$ , $W_{down}$ matrices, and Attn means to apply LoRA to $W_{q}$, $W_{k}$, $W_{v}$, $W_{o}$ matrices. In P-tuning experiments, the number of prompt tokens is set of $n=20$. In the MLP encoder, there are 3 linear layers with ReLU activation in between. In the LSTM encoder, a bidirectional LSTM is used and the output is passed to 2 linear layers with ReLU activation in between. For all the above experiments, we used the AdamW optimizer and set the learning rate to $5e-5$, per device train batch size to 4, gradient accumulation steps to 4. For the ROME and MEMIT baselines, we used the default hyperparameter settings provided in EasyEdit, shown in Table \ref{tab:rome-parameters} and \ref{tab:memit-parameters}. 

For the knowledge modification and knowledge injection experiments in Table \ref{tab:applekg_performance}, we oversample each knowledge injection samples four times due to the limited number of training examples, as generating an update example for knowledge injection is not possible. The hyperparameter settings are kept the same as above.

\end{document}